\documentclass[runningheads,review]{llncs}
\usepackage{graphicx}
\usepackage{amssymb}
\usepackage{amsmath}
\usepackage{amsopn}
\usepackage{amsfonts}
\usepackage{booktabs}
\usepackage{wrapfig,lipsum}
\usepackage{enumitem}

\usepackage{todonotes}

\makeatletter
\newcommand{\@chapapp}{\relax}%
\makeatother

\usepackage{algorithm} 
\usepackage{algpseudocode} 

\usepackage{multirow}

\begin{document}

\newcommand{\mpara}[1]{\medskip\noindent{\bf #1}}
\newcommand{\approach}{\textsc{KEdge}}
\title{Learnt Sparsification for Interpretable Graph Neural Networks}
\author{Mandeep Rathee \and
Zijian Zhang \and
Thorben Funke \and
Megha Khosla \and
Avishek Anand}
\institute{L3S Research Center, Hannover, Lower Saxony Germany
\email{\{rathee,zzhang,tfunke,khosla,anand\}@l3s.de}}

\newcommand{\pubmed}[0]{PubMed}
\newcommand{\gat}[0]{GAT}
\newcommand{\dropedge}[0]{DropEdge}
\newcommand{\ns}[0]{NeuralSparse}
\newcommand{\basic}[0]{Basic}
\newcommand{\gcn}[0]{GCN}
\newcommand{\myciteseer}[0]{Citeseer}
\newcommand{\cora}[0]{Cora}
\newcommand{\sgc}[0]{SGC}

\newcommand{\thorben}[1]{{\color{blue}[\textbf{Thorben}: #1]}}

\newcommand{\mandeep}[1]{{\color{red}[\textbf{Mandeep}: #1]}}

\newcommand{\zijian}[1]{{\color{yellow}[\textbf{Zijian}: #1]}}
\newcommand{\hardkuma}{\texttt{HardKuma}}
\newcommand{\kuma}{\texttt{Kuma}}

\authorrunning{M. Rathee et al.}
\maketitle

\begin{abstract}
Graph neural networks (GNNs) have achieved great success on various tasks and fields that require relational modeling. 
GNNs aggregate node features using the graph structure as inductive biases resulting in flexible and powerful models.
However, GNNs remain hard to interpret as the interplay between node features and graph structure is only implicitly learned.
In this paper, we propose a novel method called \approach{} for \textit{explicitly sparsifying} the underlying graph by removing unnecessary neighbors.
Our key idea is based on a tractable method for sparsification using the \textit{Hard Kumaraswamy distribution} that can be used in conjugation with any GNN model. %
\approach{} learns edge masks in a modular fashion trained with any GNN allowing for gradient-based optimization in an end-to-end fashion.
We demonstrate through extensive experiments that our model \approach{} can prune a large proportion of the edges with only a minor effect on the test accuracy. 
Specifically, in the PubMed dataset, \approach{} learns to drop more than $80\%$ of the edges with an accuracy drop of merely $\approx 2\%$ showing that graph structure has only a small contribution in comparison to node features.
Finally, we also show that \approach{} effectively counters the over-smoothing phenomena in deep GNNs by maintaining good task performance with increasing GNN layers. 

\end{abstract}

\section{Introduction}
\label{sec:intro}

Graph neural networks (GNNs) are a powerful family of models that operate over graph-structured data and have achieved state-of-the-art performance on node and graph classification tasks. 
GNNs aggregate node feature information using the input graph structure as inductive biases resulting in flexible and powerful learning models.
Although expressive and flexible, current GNNs lack interpretability and limited attention has been paid to building interpretable GNNs in general. 
Existing approaches for building explainable GNNs tend to focus on \textit{post-hoc interpretability}, i.e., they attempt to explain predictions of a GNN mode after the model has been trained~\cite{funke2021zorro,ying2019:gnnexplainer,XGNN2020}.
However, a fundamental limitation of such post-hoc approaches is that explanations might not necessarily be accurate reflections of the model decisions~\cite{rudin2019stop}. 
Second, and more worrisome, is the problem of the evaluation of interpretability techniques due to the difficulty in gathering ground truth for evaluating an explanation~\cite{lage2019evaluation,zhang2019dissonance}.
Consequently, it is unclear if the model or the explanation is inaccurate for a given decision.

An alternate design methodology is to ensure \textit{interpretability by design} by explicitly inducing \emph{sparsification} into GNN models.
Sparsification in this context refers to selecting a subset of neighborhood nodes (for a given query node) that are employed during the aggregation operation. 
There are two major benefits for explicit sparsification in GNNs.
First, explicit sparsification is sometimes necessary in real-world graphs that might contain task-irrelevant edges. 
Hence, a lack of explicit sparsification in GNNs models might result in the aggregation of unnecessary neighborhood information and degrade generalization. 
Secondly, as we show in this work, sparsification also helps to overcome the issue of \emph{over-smoothing}~\cite{li2018deeper} in deep GNNs.
\begin{figure}
  \centering
  \includegraphics[width=.9\textwidth]{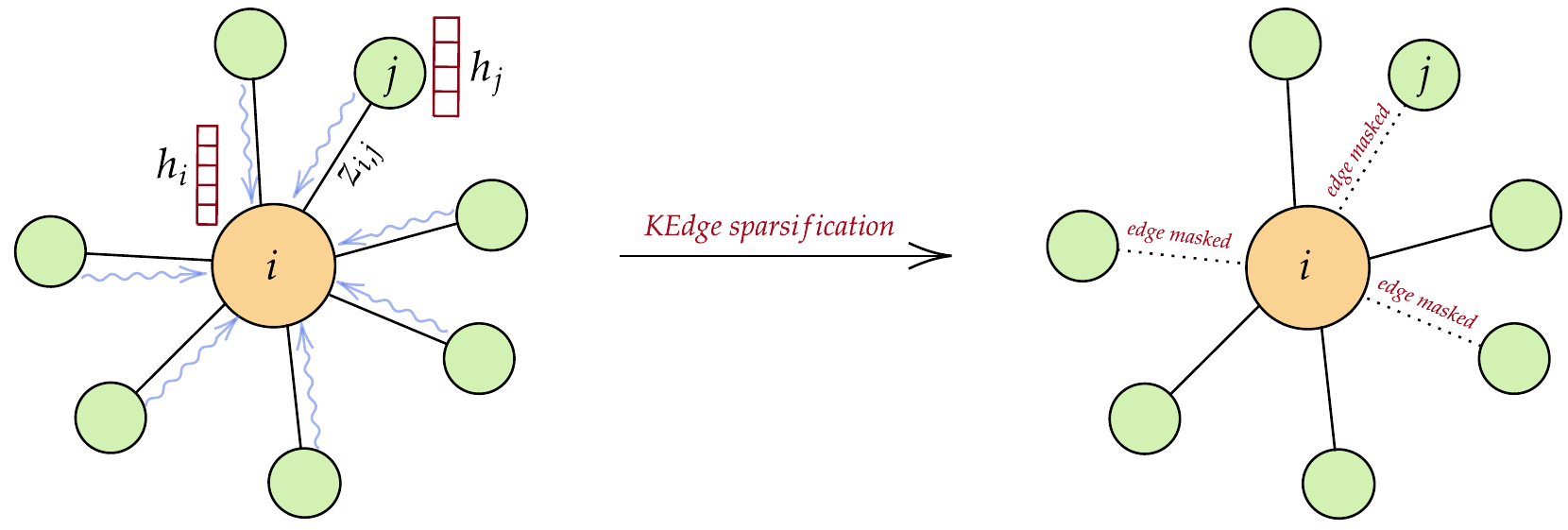}
  \caption{
  Given feature vectors of query node and its neighbor, \approach{} learns \hardkuma{} distributions for each edge in the original graph. Hence, we learn which edges can be masked without affecting performance.}
  \label{fig:1}
\end{figure}
In this work, we develop a plug-and-play \emph{interpretable by design} approach, which can be incorporated into any GNN model. 
Our approach, which we refer to as \approach{}, learns to actively sparsify the input graph towards improving model interpretability by learning to drop task-irrelevant edges.
Specifically, we build an \emph{adjacency matrix generator} layer, which is given as input the query node and its neighborhood, that selects the most important neighbors (see Figure~\ref{fig:1}), which are then used as input to the GNN layer. 
We model our neighborhood mask using the \hardkuma{} distribution~\cite{kumaraswamy1980generalized}, which exhibits a mix of discrete and continuous behavior. 
We take advantage of this particular property of \hardkuma{} distribution to achieve binary selections and gradient-based training at the same time. 
Our generator layer can be plugged either before the GNN layers or after each GNN layer giving rise to \approach{} and \approach{}-layerwise variants, respectively. 
In particular, the learned neighborhood is layer-dependent in the case of the \approach{}-layerwise variant. Intuitively, each GNN layer learns different properties of a node and might be dependent on different neighborhoods in each layer.

We demonstrate the effectiveness of \approach{} by extensive experimentation on three well-known citation networks and three state-of-the-art GNN models.
We show that \approach{} can remove a significant proportion of edges without compromising performance. 
We observe that the combination of the GNN model and dataset properties, greatly influences the ratio of removed edges by \approach{}. 
In particular, \approach{} removes $83\%$ of the edges on PubMed when using Simple Graph Convolution (SGC).
Moreover, we observe that the effect of over-smoothing on deep GNNs is significantly reduced by \approach{}-layerwise which induces a layer-specific sparser neighborhood. 

\subsection{Summary of contributions}

To summarize, we make the following contributions.
\begin{enumerate}
    \item We develop \approach{} which is based on principles of achieving interpretability via input sparsity. \approach{} is easy to incorporate into any existing GNN architecture.
    \item We develop an end-to-end trainable framework achieving binary selection and gradient-based optimization at the same time.
    \item We showcase the superiority of our approach by extensive experiments on three benchmark datasets. 
    In particular, we demonstrate our approach's effectiveness in pruning a high percentage of node neighborhoods without compromising performance.
\end{enumerate}

\section{Related Work}
\label{sec:related-work}

Graph neural networks have been popularized by the invention of graph convolutional network (GCN)~\cite{kipf2017semi} and several of its variants variants~\cite{hamilton2017inductive,duvenaud2015convolutional,atwood2016diffusion,niepert2016learning,monti2017geometric,velickovic2018graph}. The basic  GCN \cite{kipf2017semi} model compute a representation of a node via recursive aggregation and transformation of feature representations of its neighbor and trained in an end to end manner with a supervised task objective. 
Graph Attention Network \cite{velickovic2018graph} adopted attention heads, leading to weighted aggregation over node neighborhood.
With the Simple Graph Convolution (SGC), \cite{wu2019simplifying} discovered that dropping the intermediate non-linearity has a minor effect on the performance.
Others adjusted sampling approaches to reduce overfitting~\cite{rong2019dropedge,hamilton2017inductive}.

\textbf{Approaches for Graph Sparsification.} Real-world graphs are often noisy and might have a few very high degree nodes. The repeated neighborhood aggregation mechanism employed by GCNs in the presence of noise can lead to noise exaggeration by repeated error propagation. Moreover, the presence of even a small number of high degree nodes can lead to over-smoothing \cite{rong2019dropedge,oono2019graph,wu2019simplifying} of node features for deep GNNs.  To counter such effects a number of unsupervised approaches for graph sparsification \cite{calandriello2018improved,chakeri2016spectral,adhikari2017propagation,eden2018provable,voudigari2016rank,leskovec2006c,maiya2010sampling} as well as recent supervised approaches like  \dropedge~\cite{rong2019dropedge} and \ns~\cite{zheng2020robust} have been proposed. Given an input graph, the unsupervised methods extract a representative subgraph while preserving the original graph's crucial properties like its spectral properties, node-degree and distance distribution, and clustering. Supervised methods, in contrast, learn task-relevant graph structures. 
DropEdge, for example, randomly drops a specific rate of edges of the input graph during training while NeuralSparse learns k-neighbor subgraphs by selecting at most k neighbors for each node in the graph. 
Other works like LDS \cite{franceschi2019learning}, which also consider the addition of new edges or the more general problem of graph structure learning, are out of this publication's scope. We refer to~\cite{zhu2021deep} for a recent overview of approaches for graph structure learning.

\textbf{Interpretability in GNNs.} Existing works mainly focus on the post-hoc interpretability of a trained GNN model.
These approaches can be either model agnostic, such as GNNExplainer~\cite{ying2019:gnnexplainer}, XGNN~\cite{XGNN2020}, and Zorro~\cite{funke2021zorro}, or model introspective~\cite{pope2019explainability}. 
GNNExplainer learns a real-valued graph and feature mask to maximize its mutual information with GNN's predictions, while Zorro~\cite{funke2021zorro} outputs binary masks. The XGNN proposed a reinforcement learning-based graph generation approach to generate explanations for a graph's predicted class. 
Interpretable models by design that actively introduce sparsity in the input and only operate on the sparsified input have been called explain then predict models in the text domain \cite{zhang2021explain:expred}.
For GNNs, some of the models that actively introduce sparsity are \cite{zheng2020robust,louizos2017learning}.
Other approaches like~\cite{pope2019explainability} extends the gradient-based saliency map methods to GCNs and attributes them to the original model's input features.
Other works~\cite{kang2019explaine,IdahlKA19} focus on explaining unsupervised network representations, which is out of the scope of our current work.
\section{Preliminaries}

\label{sec:Notation and Prem}

\subsection{Background on Graph Neural Networks}

\textbf{Notations.} Let $G = {(V,E)}$ be an input graph with $|V|=n$ and $A\in \mathbb{R}^{n\times n}$ the adjacency matrix of the graph. $D=\operatorname{diag}(d_i)$ is the degree matrix where $d_i$ is the degree of node $v_i$. Each node $v_i$ in graph \textit{G} has d-dimensional input feature vector ${x}_{i}\in \mathbb{R}^{d}$. Let $X\in \mathbb{R}^{n\times d}$  be the input feature matrix with $X=\left[{x}_{1},{x}_{2}, \ldots, {x}_{n}\right]^{\top}$ where the $i^{th}$ row of $X$ represents input feature of node $v_i$.

In this work we focus on graph convolution network and its variants which constitute the most important class of GNNs. Basically a GCN generates a node representation by recursive aggregation and transformation of feature representations of its neighbors. 
Specifically, in each GNN layer \textit{k}, the representation ${h_i}^{(k)}$ of node $v_i$ at layer \textit{k} is obtained by aggregating information from representations of its neighbors \{${h_j}^{(k-1)}; \forall j \in \mathcal{N}_{i}$\} where $\mathcal{N}_{i}$ is set of neighbors of node $v_i$ and $h_{i}^{(0)}=x_{i}$. Overall the update step written in matrix form is
\begin{equation}
{{H}}^{(k)}=\operatorname{GNN}(A, H^{(k-1)}, W^{(k-1)})=\sigma\left( \operatorname{AGG} \{ A, {{H}}^{(k-1)} \} W^{(k-1)} \right)
\end{equation}
where $\sigma(\cdot)$ is the activation function, $H^{(0)}=X$ and $W^{(k-1)}$ is the weight matrix at $(k-1)^{th}$ layer. 

In this work, we employ three GNN models, namely GCN~\cite{kipf2017semi}, SGC~\cite{wu2019simplifying} and GAT~\cite{velickovic2018graph} which differ mainly in their aggregation function, $\operatorname{AGG}$.
For example in GCN, the aggregation is done by $\tilde{{D}}^{-\frac{1}{2}} \tilde{{A}} \tilde{{D}}^{-\frac{1}{2}}{{H}}^{(k-1)}$ matrix operation. Here $\tilde{{A}}={A}+{I}_{n}$ and $\tilde{{D}}$ is the degree matrix of $\tilde{{A}}$. 
Finally, a softmax layer is applied to the node representations at the last layer $K$ to predict the node class. 

The parameters of the model are trained by minimizing the GNN's loss function, e.g., the cross-entropy loss
\begin{equation}
\label{eq:GNNloss}
L(w,A)= -\sum_{v\in V_\text{train}} \sum_{c} \delta_{cv} \log y_{cv}, 
\end{equation}
where $c$ corresponds to a class label, $\delta_{cv} \in \{0,1\}$ indicates whether node $v$ has label $c$ or not and $y_{cv}$ is the predicted class probability for node $v$ and class $c$. Here $w$ denotes the parameters of the GNN model.

\subsection{Probabilistic Modelling of Binary Variables }
As mentioned earlier, given a graph, we are interested in learning a sparse mask over the adjacency matrix such that task-irrelevant edges are filtered out. In particular, we are interested in learning the underlying distribution of the sparse task-relevant adjacency matrix. Ideally, we would want to model the binary elements of our adjacency matrix by a suitable probability distribution such that (i) samples can be drawn efficiently and (ii) parameters of the distribution can be efficiently learned via back-propagation.

\hardkuma{}~\cite{bastings2019interpretable} is one such distribution that satisfies the above two properties. The \hardkuma{} is an extension of \textit{Kumaraswamy (\kuma{}) distribution}~\cite{kumaraswamy1980generalized}.
\begin{definition}[\kuma{} Distribution]
A random variable $Y$ is a \kuma{} distribution if its probability density function (PDF) is given by
\begin{equation}
 f_{Y}(y ; \alpha, \beta)=\alpha \beta y^{\alpha-1}\left(1-y^{\alpha}\right)^{\beta-1}, \quad y \in(0,1),
\end{equation}
where $\alpha, \beta > 0$ and has the following cumulative distribution function (CDF)
\begin{equation}
F_{Y}(y ; \alpha, \beta)=1-\left(1-y^{\alpha}\right)^{\beta}.
\end{equation}
\end{definition}

As the original \kuma{} distribution's support does not include $0$ and $1$, the Stretched \kuma{} distribution is defined as follows which has the support in $(\ell,r)$ and samples are stretched using $\ell+(r-\ell)y$. 
\begin{definition}[Stretched \kuma{} Distribution]
Let $F_{Y}(y ; \alpha, \beta)$ be the CDF of the \kuma{} distributed random variable $Y\sim \kuma{}(\alpha,\beta)$. A random variable $T$ is said to have a Stretched \kuma{} distribution with support $(\ell,r)$ if its CDF is given by
\begin{equation}
F_{T}(t ; \alpha, \beta,\ell,r)= F_{Y}\left(\frac{t-\ell}{r-\ell}; \alpha, \beta\right), \quad t\in (\ell,r),
\end{equation}
 where $\ell<0$ and $r>1$ and its PDF is given by
\begin{equation}
 f_{T}(t ; \alpha, \beta, \ell, r)=f_{Y}\left(\frac{t-\ell}{r-\ell} ; \alpha, \beta \right) \frac{1}{(r-\ell)},
\end{equation}
where $r-\ell>0$.
\end{definition}
To retrieve a sample from \hardkuma{} distribution, we sample a point $t$ from the Stretched \kuma{} distribution and apply hard sigmoid, i.e., $z = \min(1, \max(0, t))$. The likelihood for sampling $0$ is equal to sampling any point $t < 0$ for the underlying Stretched \kuma{}. Similarly, the likelihood of sampling $1$ is equivalent to sampling any $t>1$. The probability at $z=0$ and $z=1$ is thus defined in closed form as below.
\begin{definition}[\hardkuma{} Distribution~\cite{bastings2019interpretable}]
Let $\alpha, \beta >0$, $\ell<0$, and $r>1$. Then the PDF of a \hardkuma{} distribution is given by
\begin{equation}
    f_{HK}(z; \alpha, \beta, \ell, r) = 
    \begin{cases}
    F_{Y}\left(\frac{-\ell}{r-\ell} ; \alpha, \beta\right), &\text{ for } z=0,\\
    1-F_{Y}\left(\frac{1-\ell}{r-\ell} ; \alpha, \beta\right), &\text{ for } z=1,\\
     f_{T}(h ; \alpha, \beta, \ell, r), &\text{ for } z\in (0,1).
    \end{cases}
\end{equation}
\end{definition}
We make the following observations about the samples drawn from the \hardkuma{} distribution which we are useful for understanding the rational of our approach in Section~\ref{sec:Our Method}.
\begin{itemize}
    \item \textit{Observation 1:} Given that uniform random variables can be efficiently sampled from a single random source, a continuous variable $z\in [0,1]$ can be efficiently sampled from the \hardkuma{} distribution as follows
\begin{equation}
\label{eq:z}
z=\mathtt{HardKuma}(u,\alpha,\beta,\ell,r)=
\min (1, \max (0, \ell+(r-\ell)\left(1-(1-u)^{1 / \beta}\right)^{1/\alpha})),
\end{equation}
where $u$ is a uniform random variable, $\ell$ and $r$ are constants and $\alpha, \beta$ are parameters of the distribution. This re-parameterization is differentiable \textit{almost everywhere}. For derivatives we refer to~\cite{bastings2019interpretable}.
\item \textit{Observation 2}: The gradient of $z$ with respect to parameters $\alpha$ and $\beta$ exists and can be computed efficiently. 
\item \textit{Observation 3}: The expected value of \kuma{} distribution with shape parameters $\alpha$ and $\beta$ is given by
\begin{equation}
    \frac{\beta\Gamma(1+\frac{1}{\alpha})}{\Gamma(1+\beta+\frac{1}{\alpha})},
    \label{eq:kumamean}
\end{equation}
where $\Gamma(\cdot)$ is the gamma function. We will need this analytical estimate in a special case of inference in our proposed model.
\end{itemize}

\section{Our Approach}
\label{sec:Our Method}
Given an input graph $G=(V, E)$, we are interested in learning sparser connections between data points while simultaneously training the GNN's parameters.
Removing noisy and task-irrelevant edges improves task performance and increases interpretability as the class decision can now be attributed to a small subset of the neighborhood. 
Intuitively, we are interested in generating a binary mask $Z$ for the adjacency matrix such that the modified loss $L(w, A_{sprs})$ using the sparsified adjacency matrix $A_{sprs}=A\odot Z$ is minimized.

We propose to model elements of the mask $Z$ as samples from a continuous \hardkuma{} distribution which allows gradient-based optimization.
Furthermore, our choice avoids  resorting to strategies such as \textsf{REINFORCE}~\cite{williams1992simple}, which is known to show high variance, or \textit{straight-through estimation} (STE), which might lead to biased estimates~\cite{franceschi2019learning}. 
As already described in Section~\ref{sec:Notation and Prem}, the re-parameterization trick in \hardkuma{} leads to a smooth approximation of a Bernoulli random variables thus allowing for gradient based optimization.  Also,~\cite{kingma2014stochastic} showed that such re-parameterization approaches generally lead to a decrease in variance compared to approaches like \textsf{REINFORCE}.

In the next section, we describe our approach's main ingredient, i.e., the \emph{adjacency matrix generator}, which generates the sparsified adjacency matrix. We note that due to the re-parameterization trick, we can jointly learn the parameters of the generator network together with GNN model parameters. We refer to our approach as \approach{} and provide the pseudo-code of our training in Algorithm~\ref{alg:full}.

\begin{figure}
    \centering
    \includegraphics[width=0.9\textwidth]{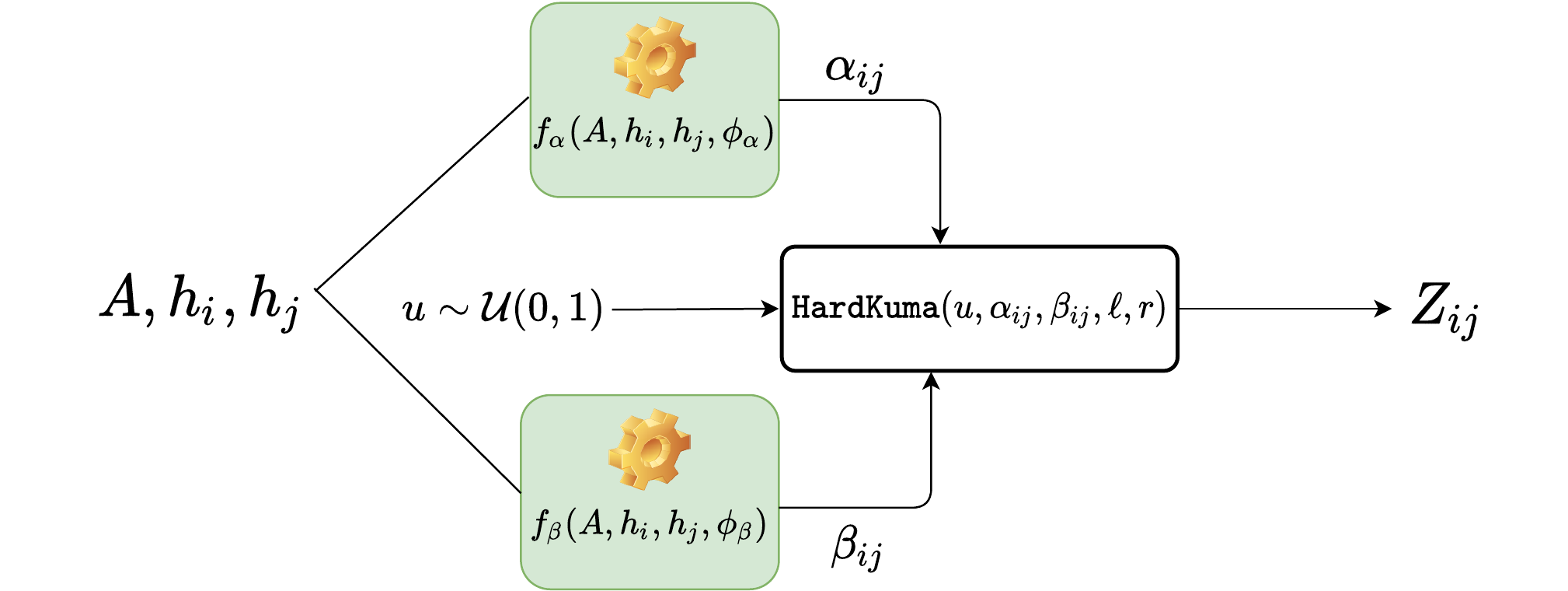}
    \caption{
    Given adjacency matrix $A$ and features of node $v_i$ and $v_i$, the neural networks learn shape-parameters $\alpha_{ij}$ and $\beta_{ij}$ for \hardkuma{}. To get mask $Z_{ij}$, $\mathtt{HardKuma}$ is applied with the learned shape-parameters to a uniform random variable $u\sim \mathcal{U}(0,1)$, where $\ell$ and $r$ are support parameters.
    }
    \label{fig:kedge}
\end{figure}

\subsection{Adjacency Matrix Generator}
Our generator network takes as input the original graph and the feature matrix and outputs a continuous mask $Z \in [0,1]^{n\times n}$. 
We model each element $Z_{ij}$ as a continuous sample from the \hardkuma{} distribution. In particular for an edge $(i,j)$ the corresponding mask $Z_{ij}$ is a  sample, see also Eq.~\eqref{eq:z}, from the \hardkuma{} distribution given by
\begin{align}
Z_{ij}& =\mathtt{HardKuma}(u,\alpha_{ij},\beta_{ij},\ell,r)\nonumber\\
&=\min \left(1, \max \left(0, \ell+(r-\ell)\left(1-(1-u)^{1 / \beta_{ij}}\right)^{1/\alpha_{ij}}\right)\right),
\label{eq:Z}
\end{align}
where $u\in(0,1)$ is a uniform random variable, $\ell$ and $r$ are the stretched lower and upper bounds of the \hardkuma{} distribution,
$\alpha_{ij}$ and $\beta_{ij}$ are parametrized using two neural networks.  
The parameters $\alpha_{ij}/\beta_{ij}$ are calculated using an attention mechanism
\begin{equation}
    \alpha_{i j}/\beta_{ij}=\frac{\exp \left(\texttt{LeakyReLU}\left({\mathrm{\theta_{\alpha/\beta}}}^{T}\left[\mathrm{W}_{\alpha/\beta} {h}_{i} \| \mathrm{W}_{\alpha/\beta} {h}_{j}\right]\right)\right)}{\sum_{k \in \mathcal{N}_{i}} \exp \left(\texttt{LeakyReLU}\left({\mathrm{\theta_\alpha}}^{T}\left[\mathrm{W}_{\alpha/\beta} {h}_{i} \| \mathrm{W}_{\alpha/\beta} {h}_{k}\right]\right)\right)},
    \label{eq:sampling_alpha_beta}
\end{equation}
where ${h}_{i}$ and ${h}_{j}$ are feature vectors of nodes $v_i$ and $v_j$, see Figure~\ref{fig:kedge}. 
In addition, we have the trainable parameters $\phi_{{\alpha/\beta}}=(\theta_{\alpha/\beta}, W_{\alpha/\beta})$ with $\theta_{\alpha/\beta}\in\mathbb{R}^{2d'}$ the attention mechanism, $W_{\alpha/\beta}\in \mathbb{R}^{d'\times d}$ the projection, and $d'$ the dimension of the attention mechanism.
$||$ is the concatenation operation and we apply the LeakyReLU nonlinearity (with negative input slope $\lambda= 0.2$).
The masked adjacency is then used as input to the GNN model.
Algorithm~\ref{alg:amg} summarizes the generation of the sparsified adjacency matrix. 

\begin{algorithm}
	\caption{Adjacency Matrix Generator: \textsc{Amg}($A, H, \phi_\alpha, \phi_\beta$)} 
	\begin{algorithmic}[1]
	\State \textbf{Input:} adjacency matrix $A$, features representation $H$, learnable parameters $\phi_\alpha=(\theta_\alpha, W_\alpha)$, $\phi_\beta=(\theta_\beta, W_\beta)$
	\State \textbf{Output: } Sparsified adjacency matrix $(A_{sprs})$
	\State \textbf{Hyperparameters:} $\ell$ and $r$ the support parameters for \hardkuma{}, $d'$ the dimension of the attention mechanism
	
	\State  For each edge in $G$ calculate $\alpha_{i j}$ and $\beta_{ij}$ based on $H, \phi_\alpha$ and $\phi_\beta$ using Eq. (\ref{eq:sampling_alpha_beta})
	\State Calculate $Z$ by applying Eq. (\ref{eq:z}) to uniform variables $u_{ij}\sim \mathcal{U}(0,1)$
	 \State \textbf{return } $A_{sprs}=A\odot Z$
	\end{algorithmic} 
	\label{alg:amg}
\end{algorithm}

\begin{figure}
  \centering
  \includegraphics[width=0.3\textwidth]{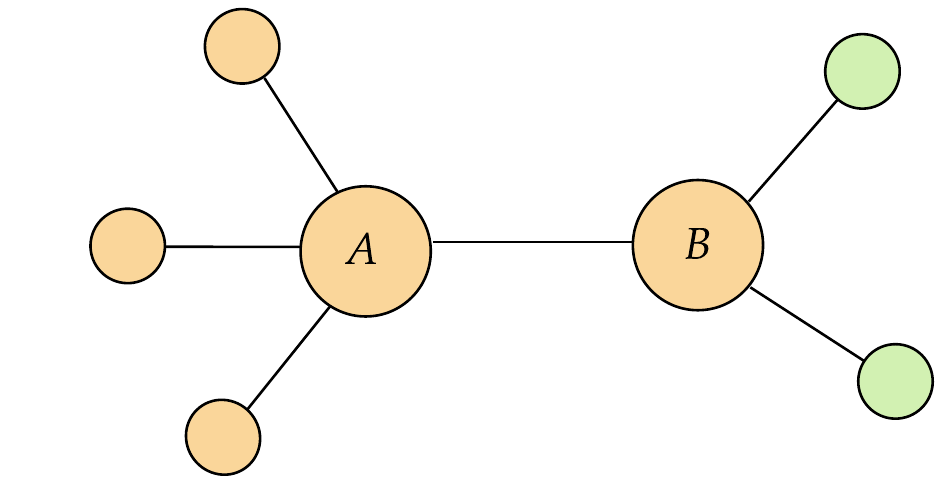}
  \caption{
  An example graph where asymmetry in neighborhood influence is expected. Node color represents its label. To predict label of node B, neighbor A should be considered. On the other hand node B might lead to error propagation to A.
  }
  \label{fig:asym}
\end{figure}

\textbf{Why does the sparsified adjacency matrix not need to be symmetric?}
Unlike previous works \cite{franceschi2019learning}
which emphasized generating a symmetric adjacency matrix, we argue that the influence of neighbors over the node label might not always be symmetric even for undirected graphs. Consider for example Figure~\ref{fig:asym}, in which nodes A and B have the same label (depicted by the same color). Moreover, all other neighbors of node B except A have a different label.
Now when B is the query node, the neighbor A should be considered as it positively influences the decision. In contrast to predict label for A, the model should better ignore the neighbor B. As B has larger fraction of neighbors with the opposite label, there is a higher chance of error propagation via B to A.

Our adjacency matrix generator (\textsc{Amg}) treats edges $\left(i,j \right)$ and $\left(j,i\right)$ independent, i.e., removal of edge $\left(i,j \right)$ is independent to the removal of edge $\left(j,i \right)$.  In particular the assymetricity in generating masks (see Eq. \eqref{eq:Z} and \eqref{eq:sampling_alpha_beta}) ensures that the generated adjacecncy matrix is not restricted to be symmetric.

\subsection{Variants of \approach{}}

\begin{figure}
  \centering
  \includegraphics[width=0.9\textwidth]{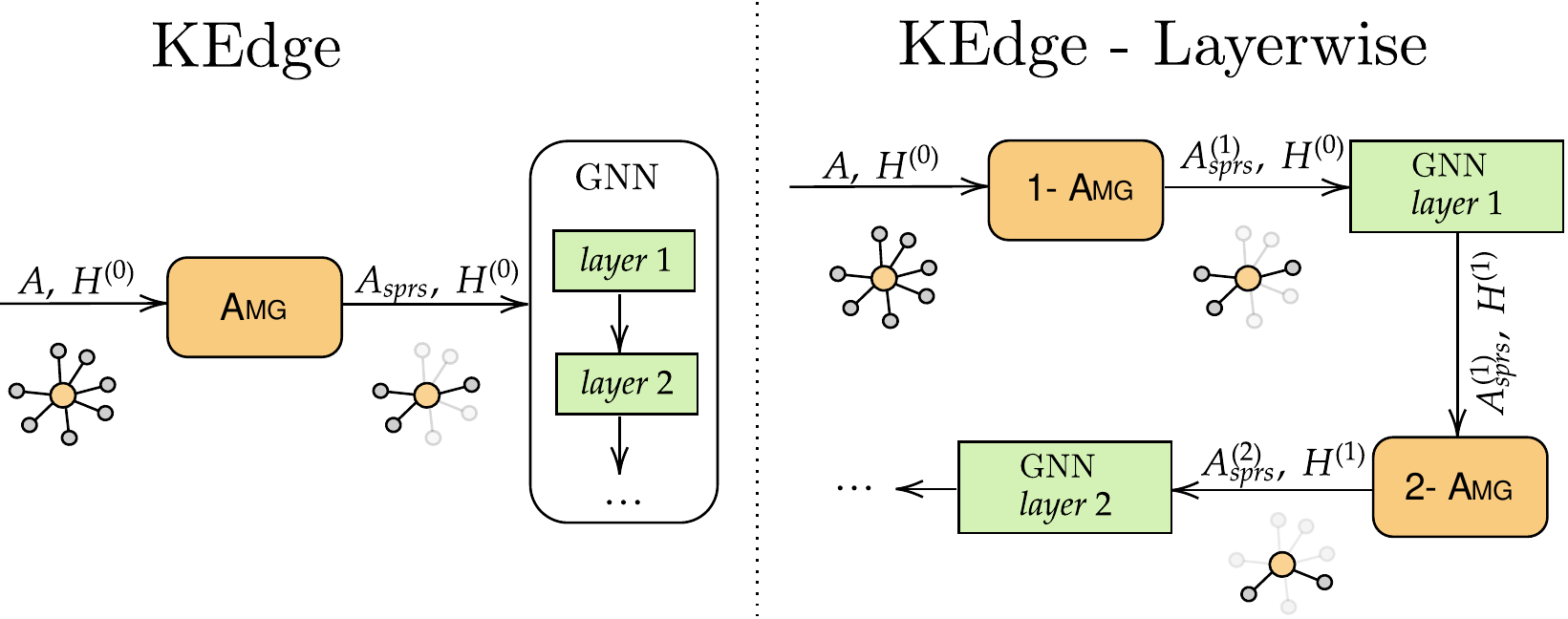}
  \caption{
  Illustration of the two \approach{} variants. In the \approach{} variant, the GNN is treated as a black-box and the adjacency is sparsified only once. In the \approach{}-layerwise variant, the adjacency $A$ is sparsified further in each layer. Note that in \approach{}-layerwise aggregation might be performed over different set of neighborhood nodes in each layer.  
  }
  \label{fig:Mi_MA}
\end{figure}

We propose two variants of our approach: the \approach{} variant, in which the GNN is treated as a black-box and the graph is sparsified by applying Adjacency Matrix Generator (\textsc{Amg}) only once at the input GNN layer, and the \approach{}-layerwise variant, in which a new sparse adjacency matrix is generated corresponding to each GNN layer (see Figure \ref{fig:Mi_MA}). 

\mpara{\approach{}.}
The \approach{} variant is a specialized version of the \approach{}-layerwise, where we only apply the first sparsification before any GNN aggregation. 
Hence, we first sparsify the original input and afterwards any GNN-model can be applied to the sparsified result. The objective function for \approach{} is given as
\begin{equation}
    \min_{{\phi_\alpha},\phi_{\beta}, W } \mathbb{E}_{u\in \mathcal{U}} (L(\{W^{(k)}\}_{k=0}^{K-1},A_{sprs}^{(1)}, \phi_{\alpha}^{(0)}, \phi_{\beta}^{(0)}))+\epsilon||Z||_{0},
    \label{eq:loss_ma}
\end{equation}
where $A_{sprs}^{(1)}=\textsc{Amg}(A,H^{(0)},\phi_{\alpha}^{(0)}, \phi_{\beta}^{(0)})$ and $Z$ is the generated mask matrix. In order to explicitly enforce sparsity, we add an additional regularization objective which corresponds to minimization of the $L0$ norm of $Z$. 

\mpara{\approach{}-layerwise.}
For \approach{}-layerwise,  we generate a new adjacency matrix for each GNN layer. 
For layer $k\geq0$, we have
\begin{equation}
    A_{sprs}^{(k+1)} = A_{sprs}^{(k)} \odot Z^{(k)} =\textsc{Amg}(A_{sprs}^{(k)}, H^{(k)},\phi_\alpha^{(k)}, \phi_\beta^{(k)}),
\end{equation}
where we use the hidden representation $H^{(k)}$ of the previous GNN-layer as input to our $\textsc{AMG}$.
The GNN-layers then operate on the sparsified adjacencies
\begin{equation}
    H^{(k+1)}=\operatorname{GNN}(A_{sprs}^{(k+1)}, H^{(k)},W^{(k)}),
\end{equation}
 where $A_{sprs}^{(0)}=A$. The overall optimization problem for KEdge-layerwise is formulated as
\begin{equation}
    \min_{{\phi_\alpha},\phi_{\beta}, W } \mathbb{E}_{u\in \mathcal{U}} (L(\{W^{(k)},A_{sprs}^{(k+1)}, \phi_{\alpha}^{(k)}, \phi_{\beta}^{(k)}    \}_{k=0}^{K-1}))+\epsilon\sum_{k=1}^{K}||Z^{(k)}||_{0},
    \label{eq:loss_mi}
\end{equation} 
where $Z^{(1)},\dots, Z^{(K)}\in[0,1]^{n\times n}$ are \hardkuma{} samples.
\subsection{Training and Inference}
\label{subsec:Train and test}
\mpara{Training.} Thanks to the re-parameterization of our binary variables using continuous \hardkuma{} distribution, we train the parameters of the GNN model as well as that of the adjacency matrix generator jointly. The resulting optimization problems are given by Eq.~\eqref{eq:loss_mi} and Eq.~\eqref{eq:loss_ma}.
We use stochastic gradient descent to minimize the loss over the training data. 
The pseudo-code of our main algorithm is given by Algorithm~\ref{alg:full}.

\begin{algorithm}
	\caption{KEdge-layerwise Training} 
	\begin{algorithmic}[1]
	\State \textbf{Input:} adjacency matrix $A=A_{sprs}^{(0)}$, node features $X=H^{(0)}$, ground-truth labels Y; learnable parameters $\phi_{\alpha}^{(.)}=(\theta_{\alpha}^{(.)}, W_{\alpha}^{(.)})$, $\phi_{\beta}^{(.)}=(\theta_{\beta}^{(.)}, W_{\beta}^{(.)})$   for \textsc{Amg}, and $W^{(.)}$ for GNN
	\State \textbf{Output:} Trained sparsified GNN
	\State \textbf{Hyperparameters:} parameters of optimizer, hyperparameters of \textsc{Amg} and $\operatorname{GNN}$, number of layers $K$
	\For{epochs}
    	\For {  $k\leftarrow0$ to $K-1$},
    		\State  $A_{sprs}^{(k+1)}\leftarrow \textsc{Amg}(A_{sprs}^{(k)}$, $H^{(k)}$, $\phi_\alpha^{(k)}$, $\phi_\beta^{(k)}$)
    
    		\State $H^{(k+1)}\leftarrow GNN(A_{sprs}^{(k+1)}$ ,$H^{(k)}$, $W^{(k)}$)
    	\EndFor
    	\State \^{Y} $\leftarrow$ prediction with $H^{(K)}$
    	\State Compute loss using $\hat{Y}$, i.e., Eq.~\eqref{eq:loss_mi}
    	\State Apply optimizer and update parameters of GNN and \textsc{Amg} 
	\EndFor
	\end{algorithmic}
	\label{alg:full}
\end{algorithm}

\mpara{Deterministic Predictions during Inference.}
At test time, we obtain for each edge $\left(i,j\right)$ a mask $z_{ij}$ based on the most likely assignment. We test using a soft or a hard mask.
(1) For \textbf{SoftMask}, we use \textit{arg-max} of the binary and continuous mask, i.e., $Z_{ij}=0$, $Z_{ij}=1$ or $0<Z_{ij}<1$. If continuous mask ($0<Z_{ij}<1$) is more likely, then we use the mean, see Eq.~\eqref{eq:kumamean}, of the learned \kuma{} distribution as the final mask. Otherwise, we use either $0$ or $1$, whichever is more likely. (2) For \textbf{HardMask}, we take \textit{arg-max} over the binary masks, i.e., $0$ and $1$. 

\section{Experimental Evaluation}
\label{sec:experiments}

In this section, we evaluate our method on the node classification task by answering two research questions: 

\newtheorem{RQ}{RQ}

\begin{RQ}
 What are the interpretability-performance trade-off of \approach{}?
\label{rq1}
\end{RQ}

\begin{RQ}
    To what extent does \approach{} improve the performance in \textbf{Deep} GNNs?
\label{rq3}
\end{RQ}

\subsection{Baseline Methods}
We compare our approach \approach{} against the two existing baseline methods for sparsification of GNNs -- \dropedge{}~\cite{rong2019dropedge} and \ns{}~\cite{zheng2020robust}. 
For these baselines, we use the default hyper-parameters as stated in the original publications.
In addition to the performance of  sparsified GNNs, we also report the performance of all GNNs without any sparsification, denoted as \textbf{\basic{}}. 

\mpara{\ns{}~\cite{zheng2020robust}.} NeuralSparse learns to select task-dependent edges by getting signals from the downstream task.  Given a hyper-parameter $k$, NeuralSparse samples k-neighbors subgraphs, which are given to the GNN as input. The process of sparsification and learning representation by the GNN is done simultaneously.

\mpara{\dropedge{}~\cite{rong2019dropedge}.} DropEdge randomly drops a fixed portion of edges from the graph before feeding it to the GNN. The process of dropping edges is uniformly, only determined by ratio, which is a hyperparameter, and only applied during training.

\subsection{Datasets \& Evaluation Methodologies}
\mpara{Datasets.} We use three benchmark graphs datasets \cora{}, \myciteseer{} and \pubmed{}
in our experiments and use the default configuration of training and test splits according to~\cite{yang2016revisiting}. 
Table~\ref{Table1} gives an overview of these datasets.
Each dataset is a citation network, where documents are represented as nodes and edges are citations. 
Nodes are represented by feature vectors and labels. 

\begin{table}
\centering
\caption{General statistics of the used datasets.}
{\begin{tabular}{crrrr}
\toprule
Dataset & Nodes & Edges & Features & Classes   \\
\midrule
Cora & 2,708 & 5,429 & 1,433 & 7\\
Citeseer & 3,327 & 4,732 &  3,703 & 6\\
PubMed & 19,717 & 44,338 & 500 & 3 \\
\bottomrule
\end{tabular}}
\label{Table1}
\end{table}
\mpara{GNN Models \& Evaluation Measures.}
We use three well-known GNNs as base models, including Graph Convolutional Network (GCN)~\cite{kipf2017semi}, Simple Graph Convolution (SGC)~\cite{wu2019simplifying} and Graph Attention Network (GAT)~\cite{velickovic2018graph} on the node classification on the aforementioned datasets. 
Since SGC first aggregates the neighbor features and then applies a single projection on the aggregates, we cannot apply our \approach{}-layerwise variant.
We report for each GNN model and sparsification method the test accuracy and the percentage of dropped edges. 
The percentage of dropped edges is calculated based on the original graph's adjacency and the last sparsified adjacency matrix. 
Hence, for \approach{}, we compare $A$ and $A_\text{sprs}$ and for \approach{}-layerwise for a 2-layer GNN, we compare $A$ and $A^{(2)}_\text{sprs}$.
Only those edges with a mask value of $0$ are counted as dropped edges for our soft-mask variant. 

\mpara{Experimental Setup.}
To answer our research questions, we perform experiments in two settings: 
First, we evaluate all three GNN models in their standard setting as 2-layer-GNNs and compare the sparsification competitors as well as our four \approach{} variants. 
Second, we focused on one combination of a dataset and a GNN model.
For GCN and Cora, we increased the number of layers of GCN from 2 to 8 to check the effect of \approach{}-layerwise (HardMask) on over-smoothing of features.

All experiments were conducted on a server with Intel Xeon Silver 4210 CPU and an Nvidia A100 GPU. 
All GNNs have a hidden feature dimension of $16$, and we use the attention mechanism $d'=16$ in the adjacency matrix generator.
Our implementation is based on PyTorch and optimizes with Adam.
For more details, see the implementation of \approach{}\footnote{https://github.com/Mandeep-Rathee/KEdge}.

\subsection{Interpretability-Performance Tradeoffs}

\begin{table}[!h]
\centering
\caption{Performance with respect to test accuracy and percentage of removed edges at inference time for semi-supervised tasks on three datasets for all sparsification methods, each with three GNNs. 
}
\label{tab:acc_diff_methods}
\begin{tabular}{cccccccc}
\toprule
GNN Model & Method  & \multicolumn{2}{c}{\cora} & \multicolumn{2}{c}{\myciteseer} & \multicolumn{2}{c}{\pubmed}\\
  &   & Acc. & \% Rem. & Acc. & \% Rem. & Acc. & \% Rem. \\
\midrule
\multirow{7}{*}{\gcn} & \basic   & 80.9 & 0 & 70.3  &  0 & 78.8& 0 \\
    & \dropedge & 79.3 & 0 & 69.9   &  0 & 77.9 & 0 \\
             & \ns & 81.7 & 20  & 68.8  &  30  & 78.4 & 48\\
             \cmidrule(r){2-8}
   &{\approach} (SoftMask)& 80.1 & 4 & 72.3 & 2  & 74.6 & 34\\
    &{\approach} (HardMask)& 79.1 & 22 & 69.8 & 16   &72.3 & 83 \\
    &{\approach}-layerwise(SoftMask)& 79.0 & 17 &   70.8 & 3  &77.5 & 54\\
   &{\approach}-layerwise(HardMask)& 76.5 &44  &  68.2 & 18  & 75.6 & 78 \\
\midrule
\multirow{7}{*}{ \sgc} & \basic   &  80.8 & 0 & 72.6 &     0 & 75.6 & 0\\
        & \dropedge& 80.8 & 0 & 72.7 & 0    & 76.2 & 0\\
             & \ns & 79.7 & 20 & 72.8 &  30  & 76.6 & 48 \\
             \cmidrule(r){2-8}
    &{\approach} (SoftMask)& 78.7 & 28 & 70.6 & 33  & 74.4 & 83\\
    &{\approach} (HardMask)& 76.3 & 42 & 67.2 & 50   & 72.0 & 92 \\
    &{\approach}-layerwise (SoftMask)& \footnotesize n.a. & \footnotesize n.a. &   \footnotesize n.a.&   \footnotesize n.a. &   \footnotesize n.a. & \footnotesize n.a. \\
   &{\approach}-layerwise (HardMask)& \footnotesize n.a. & \footnotesize n.a. &  \footnotesize n.a. & \footnotesize n.a. & \footnotesize n.a.&   \footnotesize n.a. \\    
\midrule
\multirow{7}{*}{ \gat} & \basic   & 80.9 & 0 & 70.4 &   0  & 78.3& 0 \\
        &\dropedge & 80.6 & 0 & 71.2 & 0    & 78.5 & 0 \\
    & \ns          & 80.7 & 20 & 70.3  & 30   & 77.1 & 48 \\
    \cmidrule(r){2-8}
  &{\approach} (SoftMask)& 79.8 & 3 & 70.9&  3  & 78.3 & 1\\
    &{\approach} (HardMask)& 78.5 & 8 & 67.5 & 17   & 76.9& 36\\
    &{\approach}-layerwise (SoftMask)& 81.9 & 3 &   71.3 & 3  &78.0 & 1\\
   &{\approach}-layerwise (HardMask)& 79.3 & 8 &  67.9&  16  & 76.9 & 36\\
   
\bottomrule
\end{tabular}
\end{table}

First, towards answering \textbf{RQ~\ref{rq1}}, we want to study the effect of the sparsification or increasing interpretability on the performance measured by test accuracy  with respect to removed edges. 
Table~\ref{tab:acc_diff_methods} shows the results for the three evaluated GNNs, the original model (Basic), and the two baselines, as well as our four variants of \approach{} for the three datasets. 
We observe that \approach{} and its variants can prune task-irrelevant edges without substantially affecting the generalization power of the GNNs.
For some cases, the sparsification induced by \approach{} even improves model performance. 
For example, for the combination of GAT and \approach{}-layerwise (SoftMask) on Cora, the performance is improved by $1.2\%$. 
GCN with \approach{} (SoftMask) on Citeseer has a $2.8\%$ improvement. 
In all cases, our SoftMask variant achieves higher accuracy but lower sparsification than our HardMask variant. 
This is to be expected, because of the information loss due to edge removals, but we note that the loss of performance is only marginal.
We also observe that having multiple sparsification steps, the \approach{}-layerwise often retrieves sparser adjacency matrices than \approach{}.

The ratio of removed edges is for \ns{} only determined by its hyperparameter $k$ and only affects nodes with a degree greater than $k$. 
From our results, we see that this GNN-independent choice can lead to inferior sparsification and performance. 
In contrast, our approach learns how much sparsification is possible from the data and applied GNN.
Especially noteworthy are our results on PubMed.
Our \approach{} (HardMask) model with SGC can drop up to $92\%$ of the edges while only decreasing test performance by $3$ to $4\%$. 
This is striking, since this result implies that to predict the label of a node in the PubMed dataset, the GNN model does not need to depend on the node's neighbors. 
The node features have sufficient information to predict the label. 
Overall, we conclude for our first research question that \approach{} and its variants 
result in high sparsification with minor reduction in performance (if any). High sparsification inturn to leads to improved interpretability.

\subsection{\approach{} for Deeper GNNs}
\label{sec:effect-of-depth}

\begin{figure}[h!]
  \centering
  \includegraphics[width=0.6\textwidth]{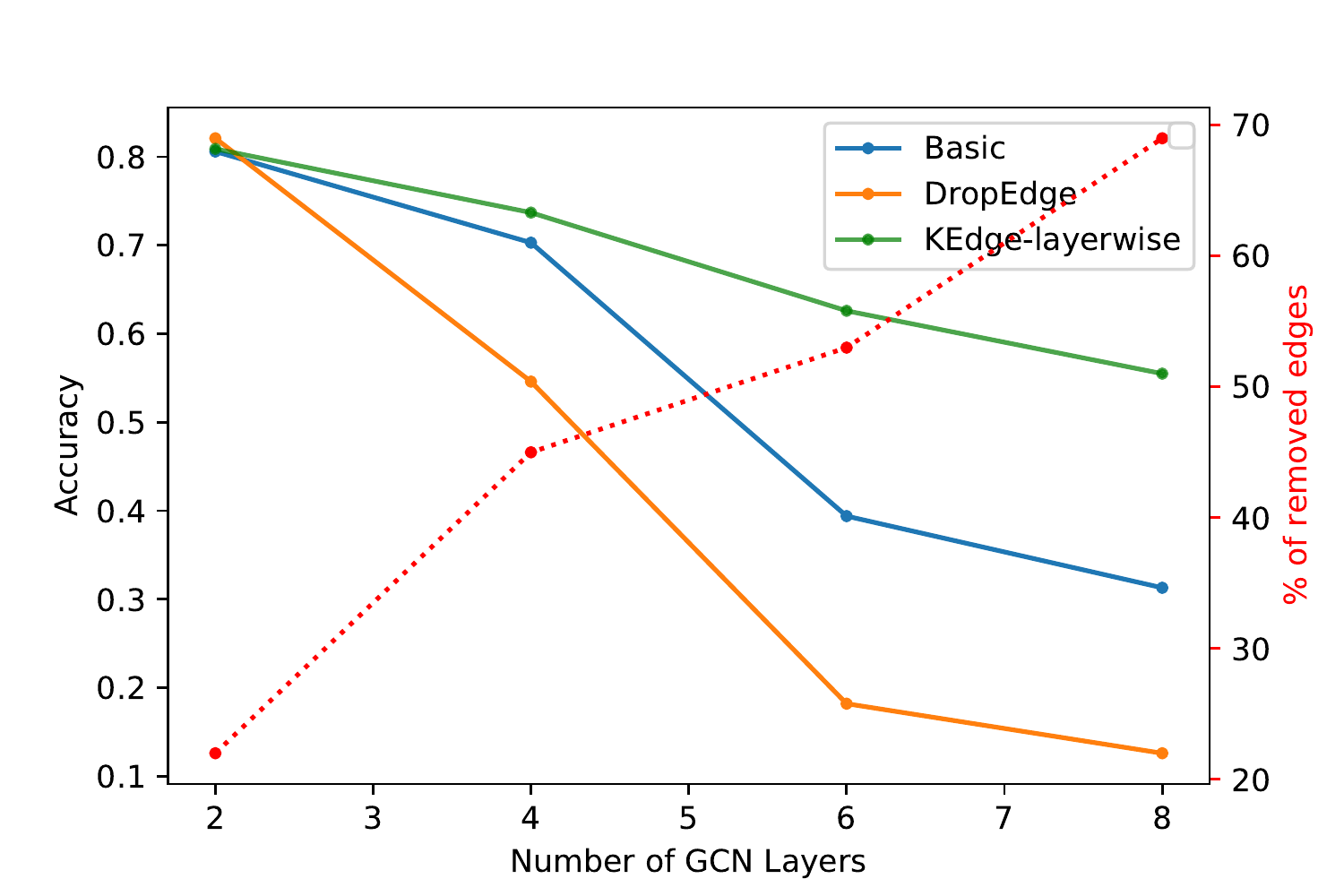}
  \caption{Effect of increasing GCN layers on accuracy. The left axis reports the achieved accuracy of the GCN (basic), DropEdge and \approach{}-layerwise (HardMask). The right axis shows the increase of removed edges by our approach (the red-dotted line). 
  }
  \label{deep_gcn}
\end{figure}

One of the common problems in GNNs is that the performance decreases with the increasing number of aggregation layers.
This behavior is commonly attributable to  over-smoothing.
Now we present the results towards answering \textbf{RQ~\ref{rq3}}, i.e., we study the effect of \approach{} with increasing number of aggregation layers in GNNs. 
Figure~\ref{deep_gcn} shows the achieved GCN's accuracy on the Cora dataset for 2, 4, 6, and 8 layers. 
DropEdge~\cite{rong2019dropedge} theoretically explains that by dropping a certain number of edges from a graph, over-smoothing can be reduced in deep GNNs. 
As Figure~\ref{deep_gcn} shows, GCN with more than four layers faces the over-smoothing effect. 
Applying DropEdge to these GCNs decreases further the performance, and hence, DropEdge cannot tackle over-smoothing. 
In contrast, our \approach{}-layerwise (HardMask) lowers the over-smoothing effect and outperforms DropEdge and original GCN. 
In conclusion, we observe that \approach{} can indeed be effectively used in GNNs with deeper layers by avoiding the problem of over-smoothing.

\section{Conclusions}
We developed a graph sparsification approach called \approach{} that can be used together with any GNN model to enhance  its generalization and interpretability. \approach{} is a task-based graph sparsification approach that learns to drop task-irrelevant edges while learning the GNN model parameters. Modeling our edge masks with HardKuma distribution allows for gradient-based optimization. In comparison to baselines, our model results in a drop of up to $83\%$ of edges without a substantial drop in model performance. Our approach, therefore, allows us to attribute any decision to a small subset of the node neighborhood, hence increasing interpretability.  
We believe that our work can be extended to ML tasks in multiple domains like Web tasks \cite{holzmann2019estimating,holzmann2017exploring}, rankings \cite{singh2020model:pcov,singh2018posthoc} and tabular data \cite{fetahu2019tablenet}.

\bibliographystyle{splncs04}
\bibliography{reference} 

\begin{thebibliography}{10}
\providecommand{\url}[1]{\texttt{#1}}
\providecommand{\urlprefix}{URL }
\providecommand{\doi}[1]{https://doi.org/#1}

\bibitem{adhikari2017propagation}
Adhikari, B., Zhang, Y., Amiri, S.E., Bharadwaj, A., Prakash, B.A.:
  Propagation-based temporal network summarization. IEEE Transactions on
  Knowledge and Data Engineering  \textbf{30}(4),  729--742 (2017)

\bibitem{atwood2016diffusion}
Atwood, J., Towsley, D.: Diffusion-convolutional neural networks. In: Advances
  in neural information processing systems. pp. 1993--2001 (2016)

\bibitem{bastings2019interpretable}
Bastings, J., Aziz, W., Titov, I.: Interpretable neural predictions with
  differentiable binary variables. arXiv preprint arXiv:1905.08160  (2019)

\bibitem{calandriello2018improved}
Calandriello, D., Lazaric, A., Koutis, I., Valko, M.: Improved large-scale
  graph learning through ridge spectral sparsification. In: International
  Conference on Machine Learning. pp. 688--697. PMLR (2018)

\bibitem{chakeri2016spectral}
Chakeri, A., Farhidzadeh, H., Hall, L.O.: Spectral sparsification in spectral
  clustering. In: 2016 23rd international conference on pattern recognition
  (icpr). pp. 2301--2306. IEEE (2016)

\bibitem{duvenaud2015convolutional}
Duvenaud, D.K., Maclaurin, D., Iparraguirre, J., Bombarell, R., Hirzel, T.,
  Aspuru-Guzik, A., Adams, R.P.: Convolutional networks on graphs for learning
  molecular fingerprints. In: Advances in neural information processing
  systems. pp. 2224--2232 (2015)

\bibitem{eden2018provable}
Eden, T., Jain, S., Pinar, A., Ron, D., Seshadhri, C.: Provable and practical
  approximations for the degree distribution using sublinear graph samples. In:
  Proceedings of the 2018 World Wide Web Conference. pp. 449--458 (2018)

\bibitem{fetahu2019tablenet}
Fetahu, B., Anand, A., Koutraki, M.: Tablenet: An approach for determining
  fine-grained relations for wikipedia tables. In: The World Wide Web
  Conference. pp. 2736--2742 (2019)

\bibitem{franceschi2019learning}
Franceschi, L., Niepert, M., Pontil, M., He, X.: Learning discrete structures
  for graph neural networks. In: International conference on machine learning.
  pp. 1972--1982. PMLR (2019)

\bibitem{funke2021zorro}
Funke, T., Khosla, M., Anand, A.: Zorro: Valid, sparse, and stable explanations
  in graph neural networks. arXiv preprint arXiv:2105.08621  (2021)

\bibitem{hamilton2017inductive}
Hamilton, W., Ying, Z., Leskovec, J.: Inductive representation learning on
  large graphs. In: Advances in neural information processing systems. pp.
  1024--1034 (2017)

\bibitem{holzmann2019estimating}
Holzmann, H., Anand, A., Khosla, M.: Estimating pagerank deviations in crawled
  graphs:hak. Applied Network Science  \textbf{4}(1),  1--22 (2019)

\bibitem{holzmann2017exploring}
Holzmann, H., Nejdl, W., Anand, A.: Exploring web archives through temporal
  anchor texts. In: Proceedings of the 2017 ACM on Web Science Conference. pp.
  289--298 (2017)

\bibitem{IdahlKA19}
Idahl, M., Khosla, M., Anand, A.: Finding interpretable concept spaces in node
  embeddings using knowledge bases. In: Machine Learning and Knowledge
  Discovery in Databases - International Workshops of {ECML} {PKDD} 2019. pp.
  229--240. Springer (2019). \doi{10.1007/978-3-030-43823-4\_20}

\bibitem{kang2019explaine}
Kang, B., Lijffijt, J., De~Bie, T.: Explaine: An approach for explaining
  network embedding-based link predictions. arXiv preprint arXiv:1904.12694
  (2019)

\bibitem{kingma2014stochastic}
Kingma, D.P., Welling, M.: Stochastic gradient vb and the variational
  auto-encoder. In: Second International Conference on Learning
  Representations, ICLR. vol.~19 (2014)

\bibitem{kipf2017semi}
Kipf, T.N., Welling, M.: Semi-supervised classification with graph
  convolutional networks. In: International Conference on Learning
  Representations (ICLR) (2017)

\bibitem{kumaraswamy1980generalized}
Kumaraswamy, P.: A generalized probability density function for double-bounded
  random processes. Journal of hydrology  \textbf{46}(1-2),  79--88 (1980)

\bibitem{lage2019evaluation}
Lage, I., Chen, E., He, J., Narayanan, M., Kim, B., Gershman, S., Doshi-Velez,
  F.: An evaluation of the human-interpretability of explanation. arXiv
  preprint arXiv:1902.00006  (2019)

\bibitem{leskovec2006c}
Leskovec, J.: C faloutsos sampling from large graphs. In: twelfth ACM SIGKDD
  International Conference of Knowledge Discovery and Data Mining (2006)

\bibitem{li2018deeper}
Li, Q., Han, Z., Wu, X.M.: Deeper insights into graph convolutional networks
  for semi-supervised learning. In: Proceedings of the AAAI Conference on
  Artificial Intelligence. vol.~32 (2018)

\bibitem{louizos2017learning}
Louizos, C., Welling, M., Kingma, D.P.: Learning sparse neural networks through
  $ l\_0 $ regularization. arXiv preprint arXiv:1712.01312  (2017)

\bibitem{maiya2010sampling}
Maiya, A.S., Berger-Wolf, T.Y.: Sampling community structure. In: Proceedings
  of the 19th international conference on World wide web. pp. 701--710 (2010)

\bibitem{monti2017geometric}
Monti, F., Boscaini, D., Masci, J., Rodola, E., Svoboda, J., Bronstein, M.M.:
  Geometric deep learning on graphs and manifolds using mixture model cnns. In:
  Proceedings of the IEEE Conference on Computer Vision and Pattern
  Recognition. pp. 5115--5124 (2017)

\bibitem{niepert2016learning}
Niepert, M., Ahmed, M., Kutzkov, K.: Learning convolutional neural networks for
  graphs. In: International conference on machine learning. pp. 2014--2023
  (2016)

\bibitem{oono2019graph}
Oono, K., Suzuki, T.: Graph neural networks exponentially lose expressive power
  for node classification. arXiv preprint arXiv:1905.10947  (2019)

\bibitem{pope2019explainability}
Pope, P.E., Kolouri, S., Rostami, M., et~al.: Explainability methods for graph
  convolutional neural networks. In: Proc. of the Conference on Computer Vision
  and Pattern Recognition. pp. 10772--10781 (2019)

\bibitem{rong2019dropedge}
Rong, Y., Huang, W., Xu, T., Huang, J.: Dropedge: Towards deep graph
  convolutional networks on node classification. In: International Conference
  on Learning Representations (2019)

\bibitem{rudin2019stop}
Rudin, C.: Stop explaining black box machine learning models for high stakes
  decisions and use interpretable models instead. Nature Machine Intelligence
  \textbf{1}(5), ~206 (2019)

\bibitem{singh2018posthoc}
Singh, J., Anand, A.: Posthoc interpretability of learning to rank models using
  secondary training data. arXiv preprint arXiv:1806.11330  (2018)

\bibitem{singh2020model:pcov}
Singh, J., Anand, A.: Model agnostic interpretability of rankers via intent
  modelling. In: Proceedings of the 2020 Conference on Fairness,
  Accountability, and Transparency. pp. 618--628 (2020)

\bibitem{velickovic2018graph}
Veli{\v{c}}kovi{\'c}, P., Cucurull, G., Casanova, A., Romero, A., Lio, P.,
  Bengio, Y.: Graph attention networks. International Conference on Learning
  Representations  (2018)

\bibitem{voudigari2016rank}
Voudigari, E., Salamanos, N., Papageorgiou, T., Yannakoudakis, E.J.: Rank
  degree: An efficient algorithm for graph sampling. In: 2016 IEEE/ACM
  International Conference on Advances in Social Networks Analysis and Mining
  (ASONAM). pp. 120--129 (2016)

\bibitem{williams1992simple}
Williams, R.J.: Simple statistical gradient-following algorithms for
  connectionist reinforcement learning. Machine learning  \textbf{8}(3-4),
  229--256 (1992)

\bibitem{wu2019simplifying}
Wu, F., Zhang, T., Souza~Jr, A.H.d., Fifty, C., Yu, T., Weinberger, K.Q.:
  Simplifying graph convolutional networks. arXiv preprint arXiv:1902.07153
  (2019)

\bibitem{yang2016revisiting}
Yang, Z., Cohen, W.W., Salakhutdinov, R.: Revisiting semi-supervised learning
  with graph embeddings (2016)

\bibitem{ying2019:gnnexplainer}
Ying, R., Bourgeois, D., You, J., Zitnik, M., Leskovec, J.: Gnn explainer: A
  tool for post-hoc explanation of graph neural networks. arXiv:1903.03894
  (2019)

\bibitem{XGNN2020}
Yuan, H., Tang, J., Hu, X., Ji, S.: Xgnn: Towards model-level explanations of
  graph neural networks. In: KDD '20. p. 430–438. Association for Computing
  Machinery (2020)

\bibitem{zhang2021explain:expred}
Zhang, Z., Rudra, K., Anand, A.: Explain and predict, and then predict again.
  arXiv preprint arXiv:2101.04109  (2021)

\bibitem{zhang2019dissonance}
Zhang, Z., Singh, J., Gadiraju, U., Anand, A.: Dissonance between human and
  machine understanding. Proceedings of the ACM on Human-Computer Interaction
  \textbf{3}(CSCW),  1--23 (2019)

\bibitem{zheng2020robust}
Zheng, C., Zong, B., Cheng, W., Song, D., Ni, J., Yu, W., Chen, H., Wang, W.:
  Robust graph representation learning via neural sparsification. In:
  International Conference on Machine Learning. pp. 11458--11468. PMLR (2020)

\bibitem{zhu2021deep}
Zhu, Y., Xu, W., Zhang, J., Liu, Q., Wu, S., Wang, L.: Deep graph structure
  learning for robust representations: A survey. arXiv preprint
  arXiv:2103.03036  (2021)

\end{thebibliography}
\end{document}